\documentclass{article}

\usepackage{url}            
\usepackage{booktabs}       
\usepackage{amsfonts}       
\usepackage{nicefrac}       
\usepackage{microtype}      
\usepackage{float}
\usepackage{enumerate}
\usepackage{caption}
\usepackage{diagbox}
\usepackage{mathrsfs}
\usepackage{amsmath}
\usepackage{amssymb}
\PassOptionsToPackage{numbers,sort&compress}{natbib}
\usepackage{natbib}
\usepackage{graphicx}
\usepackage{url}
\PassOptionsToPackage{x11names}{xcolor}
\usepackage{xcolor}
\PassOptionsToPackage{algo2e,ruled}{algorithm2e}
\usepackage{algorithm2e}
\usepackage{jmlrutils}
\usepackage{hyperref}
\usepackage{nameref}

\newcommand{\E}{{\rm I\!E}}
\usepackage{wrapfig}

\DeclareMathOperator{\R}{\mathbb{R}}

\usepackage{pdfpages} 

\newcommand{\pr}{{\rm I\!P}}
\newcommand{\cv}{{\mbox{CV}_{loo}}}

\newtheorem{defn}{Definition}[section]

\newtheorem*{lemmat}{Lemma}
\newtheorem{claim}{Claim}[section]

\newcommand{\hh}{{\cal H}}

\newcommand{\Siz}{S_{i,z}}

\usepackage[accepted]{icml2021}

\icmltitlerunning{Distribution of Classification Margins}

\begin{document}
	
%
%

\twocolumn[
\icmltitle{Distribution of Classification Margins: \\ Are All Data Equal?}



\icmlsetsymbol{equal}{*}
\icmlsetsymbol{equal2}{$\dagger$}

\begin{icmlauthorlist}
\icmlauthor{Andrzej Banburski}{equal,to}
\icmlauthor{Fernanda De La Torre}{equal,to}
\icmlauthor{Nishka Pant}{equal2,to,goo}
\icmlauthor{Ishana Shastri}{equal2,to}
\icmlauthor{Tomaso Poggio}{to}
\end{icmlauthorlist}

\icmlaffiliation{to}{Center for Brains, Minds + Machines, MIT, MA, USA,}
\icmlaffiliation{goo}{Brown University, RI, USA}
\icmlcorrespondingauthor{Andrzej Banburski}{kappa666@mit.edu}

\icmlkeywords{Machine Learning, ICML}

\vskip 0.3in
]



\printAffiliationsAndNotice{\icmlEqualContribution} 

	\begin{abstract}

          Recent theoretical results show that gradient descent on
          deep neural networks under exponential loss functions
          locally maximizes classification margin, which is equivalent
          to minimizing the norm of the weight matrices under margin
          constraints. This property of the solution however does not
          fully characterize the generalization performance.  We
          motivate theoretically and show empirically that the area
          under the curve of the margin distribution on the training
          set is in fact a good measure of generalization.  We then
          show that, after data separation is achieved, it is possible
          to dynamically reduce the training set by more than 99\%
          without significant loss of performance.  Interestingly, the
          resulting subset of ``high capacity'' features is not
          consistent across different training runs, which is
          consistent with the theoretical claim that all training
          points should converge to the same asymptotic margin under
          SGD and in the presence of both batch normalization and
          weight decay.
	\end{abstract}
        
	\section{Introduction}
	
	The key to good predictive performance in machine learning is
	controlling the complexity of the learning algorithm. Until
	recently, there was a puzzle surrounding deep neural networks
	(DNNs): there is no obvious control of complexity -- such as
	an explicit regularization term -- in the training of
	DNNs. Recent theoretical results
	\cite{2019arXiv190507325S,DBLP:journals/corr/abs-1906-05890,PLB2020natcom,PNAS2020},
	however, suggest that a classical form of norm control
	is present in DNNs trained with gradient descent (GD)
	techniques on exponential-type losses. In particular, GD
	induces dynamics of the normalized weights which converge
	for $t \to \infty$ towards an infimum of the loss that
	corresponds to a maximum margin solution.
	
	What remains unclear, however, is the link between the minimum
        norm solutions and expected error.  In this paper, we
        numerically study the behavior of the distribution of margins
        on the training dataset as a function of time. Inspired by
        work on generalization bounds \cite{2017arXiv170608498B}, we
        provide evidence that the the area under the distribution of
        properly normalized classification margins is a a good
        approximate measure to rank different minima of a given
        network architecture.

	The intuition is that deep minima should have a margin
        distribution with a relatively small number of points with
        small margins. This in turn suggests a training algorithm that
        focuses only on the training points that contribute to the
        stability of the algorithm -- that is, on data points close to
        the separation boundary (once it has been established in the
        terminal phase of training, i.e. after hitting 0\%
        classification error during training).  We show, in fact,
        that, once separation is achieved, good test performance
        depends on improving the margin of a small number of
        datapoints, while the majority can be dropped (keeping only
        200 from the initial 50k in CIFAR10, for example). These
        results suggest that certain points in the training set may be
        more important to classification performance than others. It
        is then natural to pose the following question: can we, in
        principle, predict which data are more important? However, we
        show that the points that mostly support the dynamics are not
        consistent between different training runs due to the initial
        randomness. Moreover, it turns out that before data
        separation, there is no clear pattern to discern which
        datapoints will contribute the most. Quite interestingly, this
        result is consistent with a recent theoretical
        prediction\cite{PoggioLiao2020}: under stochastic gradient
        descent and in the presence of Batch Normalization and weight
        decay, all training points should asymptotically converge to
        the same margin and be effectively equivalent to each other.
        The randomness of which datapoints contribute the most to
        classification performance is consistent with the
        prediction. It also suggests the conjecture that in
        overparametrized models, we should expect the most important
        features learned by the network to be dependent on random
        factors such as initialization.

	\subsection{Related Work}
	
	Following the work of \cite{DBLP:journals/corr/ZhangBHRV16},
	which showed that overparametrized networks trained on
	randomized labels can achieve zero training error and expected
	error at chance level, recent papers analyze the dynamics of
	gradient descent methods. \cite{NIPS2018_8038,
		DBLP:journals/corr/abs-1811-08888, allen2018convergence,
		du2018gradient, DBLP:journals/corr/abs-1811-03804} showed
	convergence of gradient descent on overparametrized non-linear
	neural networks. Empirical work shows that sharp minima
	generalize better than flat minima, that the optimization
	process converges to those with better generalization,
	\cite{liao2018surprising, keskar_large-batch_2016} and that
	better noise stability (stability of the output with respect
	to the noise injected at the nodes of the network) correlate
	with lower generalization error \cite{chaudhari2019entropy,
		morcos2018importance, langford2002not}.  Several lines of
	research propose low complexity measures of the learned
	network to derive generalization bounds.
	Spectrally-normalized margin-based generalization bounds are
	derived in \cite{2017arXiv170608498B}, which we test here. Bounds obtained
	through a compression framework that reduces the effective
	number of parameters in the networks based on noise stability
	properties are described in \cite{2018arXiv180205296A} who
	more recently provided a sample complexity bound that is completely
	independent of the network size \cite{arora2019fine}.
		Our algorithms for dataset compression are indirectly 
	related to the data-distillation approach introduced in
	\cite{DBLP:journals/corr/abs-1811-10959} and to the  noise
	stability described in \cite{2017arXiv170608498B}. In preparing to put 
this version on arXiv, another paper dealing with reduction of training examples appeared \cite{paul2021deep}.

	\section{Theoretical motivations}
	We start with a short review of the recent theoretical findings that inspire our
	numerical investigations. 
		\subsection{Notation and Background}
	
	In this paper we assume the standard framework of supervised learning via
	Empirical Risk Minimization (ERM) algorithms for classification problems. For details see Supplementary Material or papers such
	as   \cite{Mukherjee2006}.

	{\it Deep Networks} We define a deep network with $K$ layers
	with the usual coordinate-wise scalar activation functions
	$\sigma(z):\quad \mathbf{R} \to \mathbf{R}$ as the set of
	functions
	$f(W;x) = \sigma (W^K \sigma (W^{K-1} \cdots \sigma (W^1
	x)))$, where the input is $x \in \mathbf{R}^d$, the weights
	are given by the matrices $W^k$, one per layer, with matching
	dimensions. For simplicity we consider homogenous functions, i.e. without bias terms.
	In the case of binary classification, the labels
	are $y \in\{-1,1\}$. The activation function is the ReLU
	activation.  For the network, homogeneity of the ReLU implies
	$f(W;x)=\prod_{k=1}^K \rho_k f(V_1,\cdots,V_K; x)$, where
	$W_k=\rho_k V_k$ with the matrix norm $||V_k||_p=1$ and
	$||W_k||=\rho_k$.  In the binary case, when $y_n f(x_n) >0$
	$\forall n=1,\cdots, N$ we say that the data are {\it
		separable} wrt $f \in \mathbb{F}$, that is they can all be
	correctly classified.  We define the margin of $x_n$ as
	$\eta_n = y_n f(x_n)$ and the margin of the whole dataset as
	the smallest of all margins $\eta = min_n \eta_n$,
	corresponding to a \emph{support vector} $x^*$.  For the
	multi-class case, if the prediction score is a vector
	$\{f_1(x),\ldots,f_C(x)\}$ for $C$ classes, with
	$f_{y_n}(x_n)$ the prediction for the true class, then the
	margin for $x_n$ is
	$\eta_n = f_{y_{n}}(x_n) - \max\limits_{j \neq y_n}f_{j}(x_n)$.

	{\it Dynamics \& margin maximization}
	It has been known for some time now that the norm $\rho = \prod_k \rho_k$ diverges to infinity
	as we run GD \cite{DBLP:journals/corr/abs-1906-05890, theory_III,PLB2020natcom}. This means that
	the weights $W_k$ do not converge in any meaningful sense, and it is only sensible to study the convergence of the normalized weights $V_k$.
	
	When we minimize exponential-type losses (like the exponential loss, logistic, or cross-entropy), we 
	expect that, asymptotically, the convergence to a data-separating solution only depends on data
	with the least negative exponents, i.e. the points with the smallest classification margin (the equivalent of support vectors in SVMs). Minimization of exponential-type losses then corresponds to
	the problem of maximizing the classification margin. Recent body of work has been showing that SGD biases highly over-parametrized deep networks towards  solutions that locally maximize margin \cite{DBLP:journals/corr/abs-1906-05890, theory_III,PLB2020natcom} in presence of normalization techniques (such as batch normalization).
	
	{\it Neural Collapse} Given the margin maximization results,
        the natural question one might ask is which data are the ones
        that contribute the most to the solution. As in the case of
        linear systems, the answer clearly depends on the amount of
        overparametrization.  Recent empirical observations suggest that with
        overparametrization, SGD leads to the phenomenon of
        Neural Collapse \cite{Papyan24652} after data separation is
        achieved. One of the Neural Collapse properties (NC1) says
        that within-class activations all collapse to their class
        means, implying that  the margins of
        all the training points converge to the same value. While this
        might seem unintuitive, embedding $N$ points in
        $D$-dimensional space with $D\gg N$ allows for many
        hyperplanes equidistant from all the points. A recent
        theoretical analysis predicts that NC1\cite{4625} depends on both the use of
        normalization algorithms (such as batch normalization) and L2
        regularization during SGD. The prediction  is consistent with our results in Figure
        \ref{main3}.
	
	The theory thus suggests that effectively only one
        example per class is needed to describe the decision
        boundaries of the learned model -- and that any of the
        training points could be used.  We explore in
        this work whether this prediction is correct.

	    \begin{figure}[t!]
    	\centering

    	\includegraphics[trim = 0 10 0 0, width=0.94\columnwidth,clip,draft=false,]{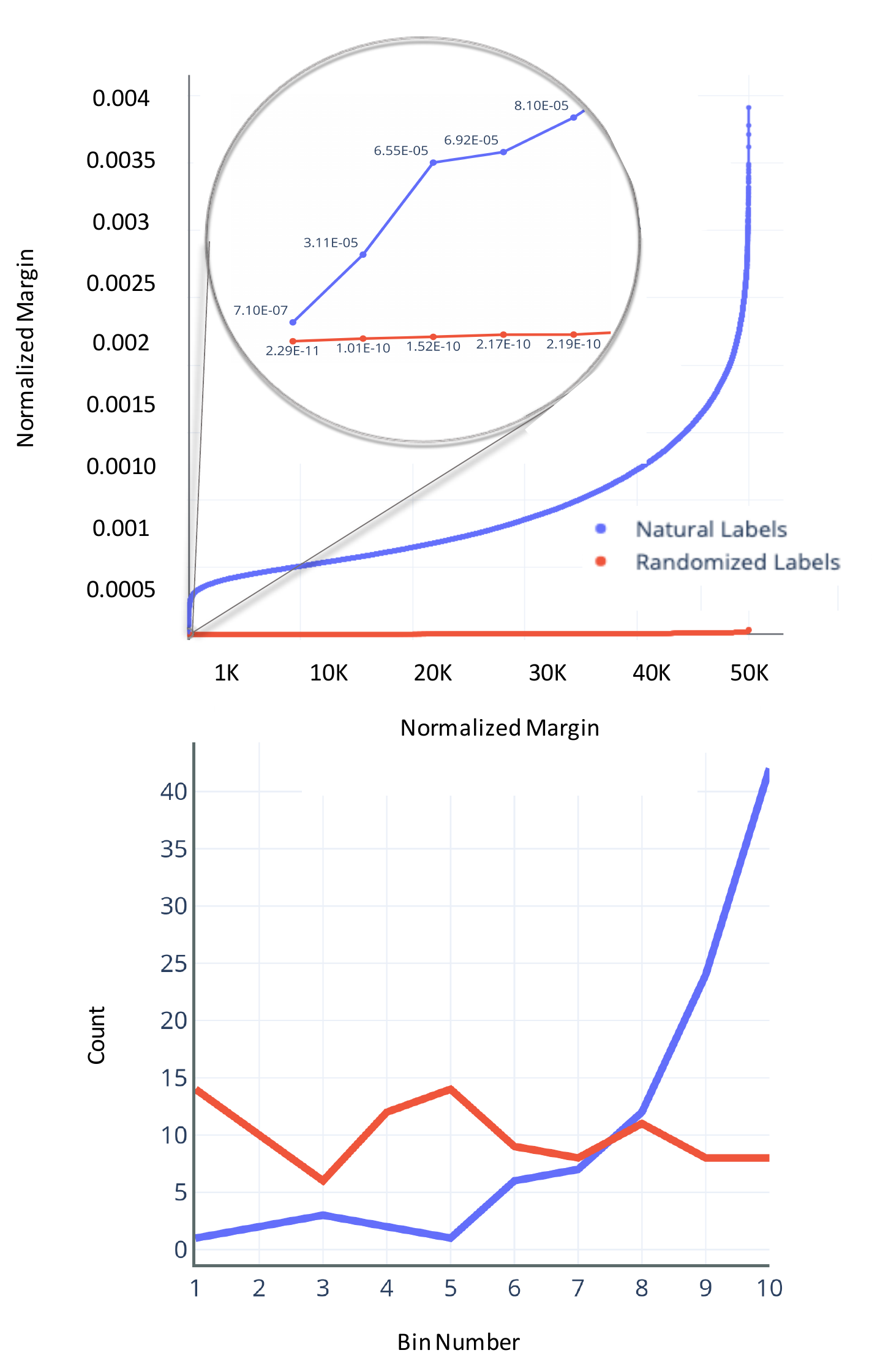}
    	\caption{\it \textbf{Natural and Random Labels - Margins} 
    		The top figure shows the margin 
			of the 50k datapoints in CIFAR10 ranked by their individual margin size for 2 convolutional networks 
			trained on either natural or randomized labels pass data separation and margin convergence. The circle enlarges the numerical values of the five datapoints with the smallest margins. 
			In the bottom figure, the range of the margin of the first $100$ datapoints (those with the smallest margin) 
			was equally divided into 10 bins with the count of data points in each bin shown. The first two bins of the 
			network trained with random labels have significantly more datapoints, while the network trained with natural labels 
			ends with less support vectors closer to each other. 
    	}\label{main1}
        \end{figure}

	\subsection{Margins, $\rho$ and expected error}
\label{MarginBounds}

Assuming that weight decay, small initialization, and batch normalization provide a bias towards a solution
with ``large'' margin, the obvious question is whether we can obtain any guarantees of good test performance. While predicting test performance purely from training behavior is challenging, we use simple bounds \cite{Bousquet2003} to predict relative performance between different minima for the same network architecture.

 A typical generalization bound
that holds with probability of at least $(1-\delta)$,
$\forall g \in \mathbb{G}$ has the form \cite{Bousquet2003}:
\begin{equation}
|L(g) -\hat{L}(g)| \leq  c_1\mathbb{R}_N(\mathbb{G}) + c_2 \sqrt
\frac{\ln(\frac{1}{\delta})}{2N},
\label{bound}
\end{equation} 
where $L(g) = \mathbf E [\ell_{\gamma}(g(x), y)]$ is the
expected loss, $\hat{L}(g) $ is the empirical loss, 
$\mathbb{R}_N(\mathbb{G})$ is the empirical Rademacher average of
the class of functions $\mathbb{G}$ measuring its complexity, and
$c_1, c_2$ are constants that reflect the Lipschitz constant of the loss function and the architecture of the network. The loss function here is  the {\it ramp
loss} $\ell_{\gamma}(g(x), y)$ defined in \cite{2017arXiv170608498B} as discounting 
predictions with margin below some arbitrary cutoff $\gamma$ (with $\ell_{0}$ being the 0-1 error, see Sup. Mat.).



We now consider two solutions with the same small training loss obtained with the
same ReLU deep network and corresponding to two different minima with
two different $\rho$s and different margins. Let us call them $g^a(x)=\rho_a f^a(x)$ and
$g^b(x)=\rho_b f^b(x)$ and let us assume that $\rho_a < \rho_b$. Using the notation of this paper, the
functions $f_a$ and $f_b$ correspond to networks with normalized weight matrices at each layer.

We now use the observation  that, because of homogeneity of the
networks, the  empirical Rademacher complexity satisfies the property
$
\mathbb{R}_N(\mathbb{G}) = \rho \mathbb{R}_N(\mathbb{F}),
$
where $\mathbb{G}$ is the space of functions of our
unnormalized networks and $\mathbb{F}$ denotes the corresponding
normalized networks.
This observation allows us to use the bound in Equation
\ref{bound} and the fact that the empirical {\it  $\hat{L_{\gamma}}$ for
both functions is the same} to write $
L_{0}(f^a)=L_{0}(F^a) \leq \hat{L_{\gamma}} +  c_1 \rho_a \mathbb{R}_N(\tilde{\mathbb{F}}) + c_2 \sqrt
\frac{\ln(\frac{1}{\delta})}{2N}$ and $L_{0}(f^b)=L_{0}(F^b) \leq \hat{L_{\gamma}}+ c_1 \rho_b \mathbb{R}_N(\tilde{\mathbb{F}}) + c_2 \sqrt
\frac{\ln(\frac{1}{\delta})}{2N}$.  The bounds have the form
\begin{equation}
L_{0}(f^a)\leq  A \rho_a  + \epsilon
\qquad \textnormal{and} \qquad
L_{0}(f^b)\leq  A \rho_b  + \epsilon
\end{equation}

{\it Thus the bound for the expected
  error $L_{0}(f^a)$ is better than the bound for $L_{0}(f^b)$.} 
Similar results can be obtained taking into account different
$\hat{L}(f)$ for the normalized $f^a$ and $f^b$ under different
$\gamma$ in Equation \ref{bound}.


Can these bounds be meaningful in practice? The  solutions $a$ and $b$ achieve the same training loss,
 which means that they must both have different norms $\rho$ and different distributions of classification margins. In what follows, we show empirically that indeed we can effectively predict the relative generalization 
performance using the information of the distribution of classification margins on the  training set.

	\section{Experimental methods}
	\label{expmethods}
		In most of the numerical experiments, we used a 5-layer neural network implemented in PyTorch and trained on the CIFAR10 or CIFAR2 (cars and birds from CIFAR10) datasets using either SGD or full GD 
	with cross-entropy loss. The network has 4 convolutional layers (filter size 3 $\times$ 3, stride 2) 
	and one fully-connected layer. All convolutional layers are followed by a ReLU activation, and for some experiments, batch normalization. The number of  channels in hidden 
	layers are 16, 32, 64, 128 respectively. In total, the network has $273,546$ parameters. The dataset was not shuffled. The learning rate was constant  and set to 0.01, with momentum set to 0.9 unless otherwise stated. Our test performance is not state of the art, since we wanted to perform neither data augmentation 
	nor any explicit regularization to match the theoretical setting. These results however extend to networks with state of the art performance, see Table \ref{compression-table} for results on DenseNet-BC and more in Sup. Mat. 
	
	In the absence of batch normalization, the margin distribution of each network is normalized by $\rho$, the product of convolutional layer norms. For networks with batch-normalization, the margin distribution was normalized by the product of the batch-normalization layer norms and the norm of the last fully-connected layer. 
	
	\section{Margin distribution}
	
	The recent Neural Collapse \cite{Papyan24652} results would suggest that at convergence the margin distribution should be flat. Convergence in the margin however is known to be very slow \cite{2017arXiv171010345S}. In this section we experimentally study the shape of the distribution of margins on the whole training dataset and then go onto using it to predict generalization performance.
	
	\subsection{Natural vs random labels}
	
		We ran numerical experiments to find the relation between the margin, stability and generalization gap for two convolutional neural networks. One was trained with natural
	labels and the second one trained with randomized labels, an idea explored in \cite{DBLP:journals/corr/ZhangBHRV16, liao2018surprising}.

	In Figure \ref{main1}, we took both networks after data separation and close to margin convergence and extracted the margins for each data point, which allowed us to sort them according to the margin. 
	The margins for all datapoints are larger for natural labels than for randomized labels. Theory suggests that the datapoints important for margin maximization 
	should be closer to each other in the feature space for randomized labels than for natural labels, consistent with lower algorithmic stability \cite{kn2002} and chance performance. To test this, we took the 100 datapoints 
	with the smallest margin for both networks, divided the margin range into ten bins and counted the number of datapoints in each bin. The first five bins for the network trained with natural labels
	had less datapoints than for randomized labels, with the first bin only having one datapoint for natural labels while for randomized labels there were 14. This experiment supports the idea that having 
	a smaller set of datapoints with small margin (the equivalent of support vectors in SVMs) leads to both better stability and test performance.
	This should be contrasted with Neural Collapse  -- we find that the margin distribution after 200 epochs of SGD is far from flat, but rather has a few small margin datapoints and a similar number of high margin points, with a flatter middle range. 
	
	\subsection{Margin distribution and generalization performance}

	How can we use the information about the margin distribution to predict generalization performance? In \cite{liao2018surprising}, it was shown that the training loss evaluated on the normalized deep network allows for a reasonable prediction of test loss, which conforms to arguments from Section \ref{MarginBounds}. It is natural to ask then whether the smallest normalized margin or a simple function of all the margins is a potentially finer measure. 
	
	\begin{figure}[t]
	\centering
	\includegraphics[trim = 0 0 0 31, width=0.93\columnwidth,clip,draft=false,]{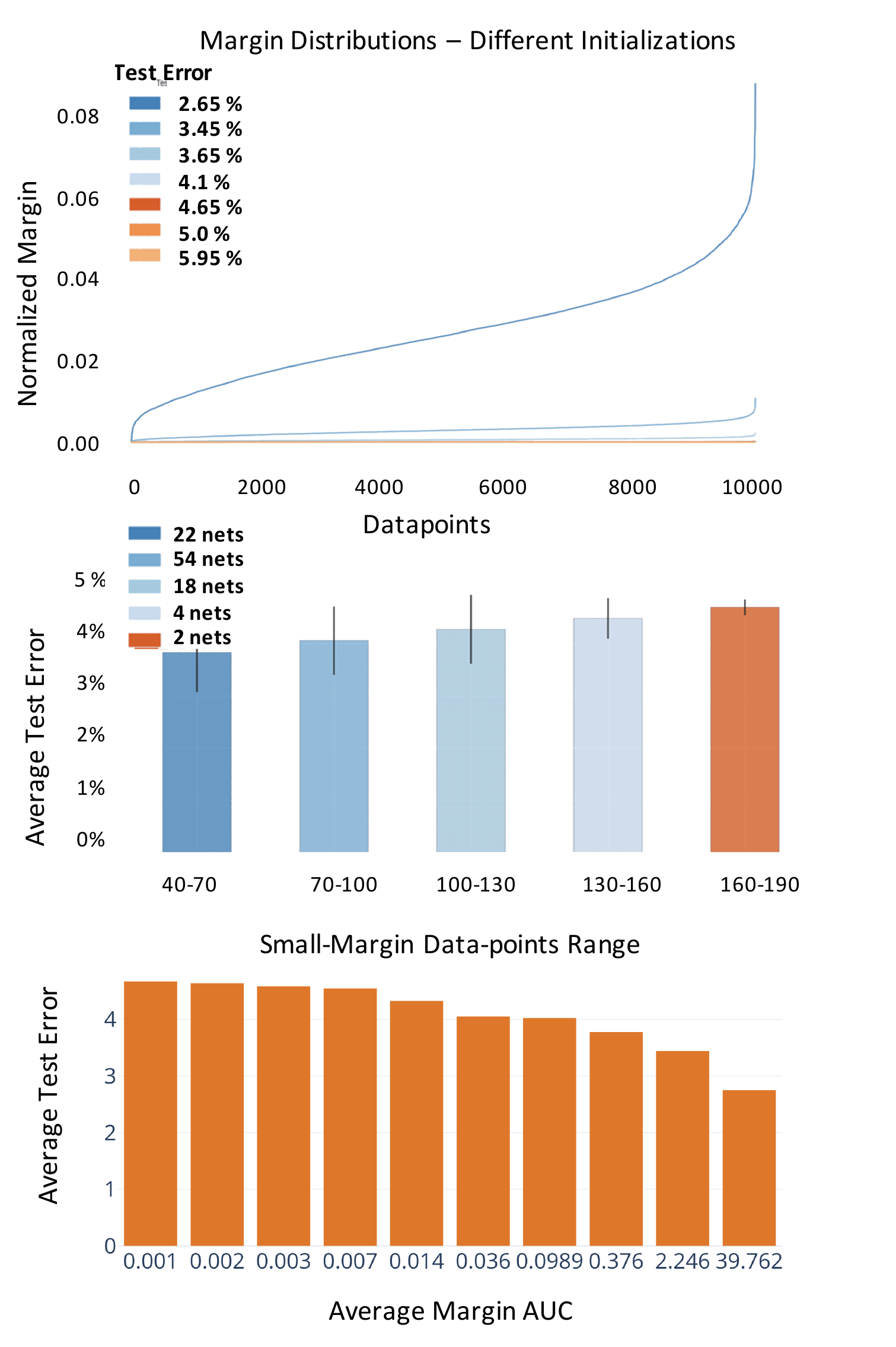}
	\caption{\it \textbf{Different Minima - Margin Distributions} 
		100 conv nets (as in Section \ref{expmethods}) were initialized with varying standard deviation (from 0.01 - 0.05) so that they  converge to different test errors. The top figure shows the margin distribution of 7 representative networks for each test performance. The middle image shows all 100 networks divided into bins given their number of small margin data and the average test error of these ranges. The bottom provides evidence that the AUC of the margin distribution is a predictor of generalization performance. Here we plotted the result with a cutoff of $\gamma = 0.1$.
	}\label{main2}
    \end{figure}
	
	To probe several metrics, we ran 100 networks on a CIFAR2 classification task, where architecture and hyperparameters stayed constant across all networks but the standard deviation for random initialization was varied. This was motivated by \cite{liao2018surprising}, since we wanted to obtain networks that converge to different minima and analyze their resulting margin distributions. We found that the area under the curve (AUC) of the margin distribution is a good predictor of generalization performance as seen in the bottom of Figure \ref{main2}. Moreover, the shape of the margin distribution is also a predictor: the initial curvature of the distribution indicates how many datapoints have small margins. Our experiments show that the number of small margin data can predict the range of test error. 
	
	These results are shown in Figure \ref{main2}: on the top we can see the sorted margin distributions of 7 representative networks for each minima (there are 10,000 datapoints in CIFAR2). For smaller test errors, the normalized margin distribution contains higher values, higher curvature, and starts off with a higher slope relative to the margin distributions of large test error (larger initialization). In the middle, we counted the number of datapoints with small margins for each network (using a margin cutoff of 0.01 above the smallest margin, which corresponds to setting $\gamma = 0.01$ in the ramp loss) and calculated the average test error for networks with different ranges of small margin points. As shown, we find that larger proportion of small margin data results in higher averages of test error. On the bottom, we divided the 100 networks into sorted bins of 10 and calculated the average AUC and average test error for $\gamma = 0.1$. We see that the larger the margin AUC, the better the test performance.
	
	In Section \ref{MarginBounds} we derived bounds for two different minima and asked if these bounds could be meaningful in practice. These experiments suggests that the shape of the margin distributions and area under it can indeed effectively predict relative generalization performance.

	\subsection{Time evolution of the margin distribution}
	
\begin{figure*}[t!]\centering
		\includegraphics[trim = 0 33 10 10, width=1.0\textwidth, clip]{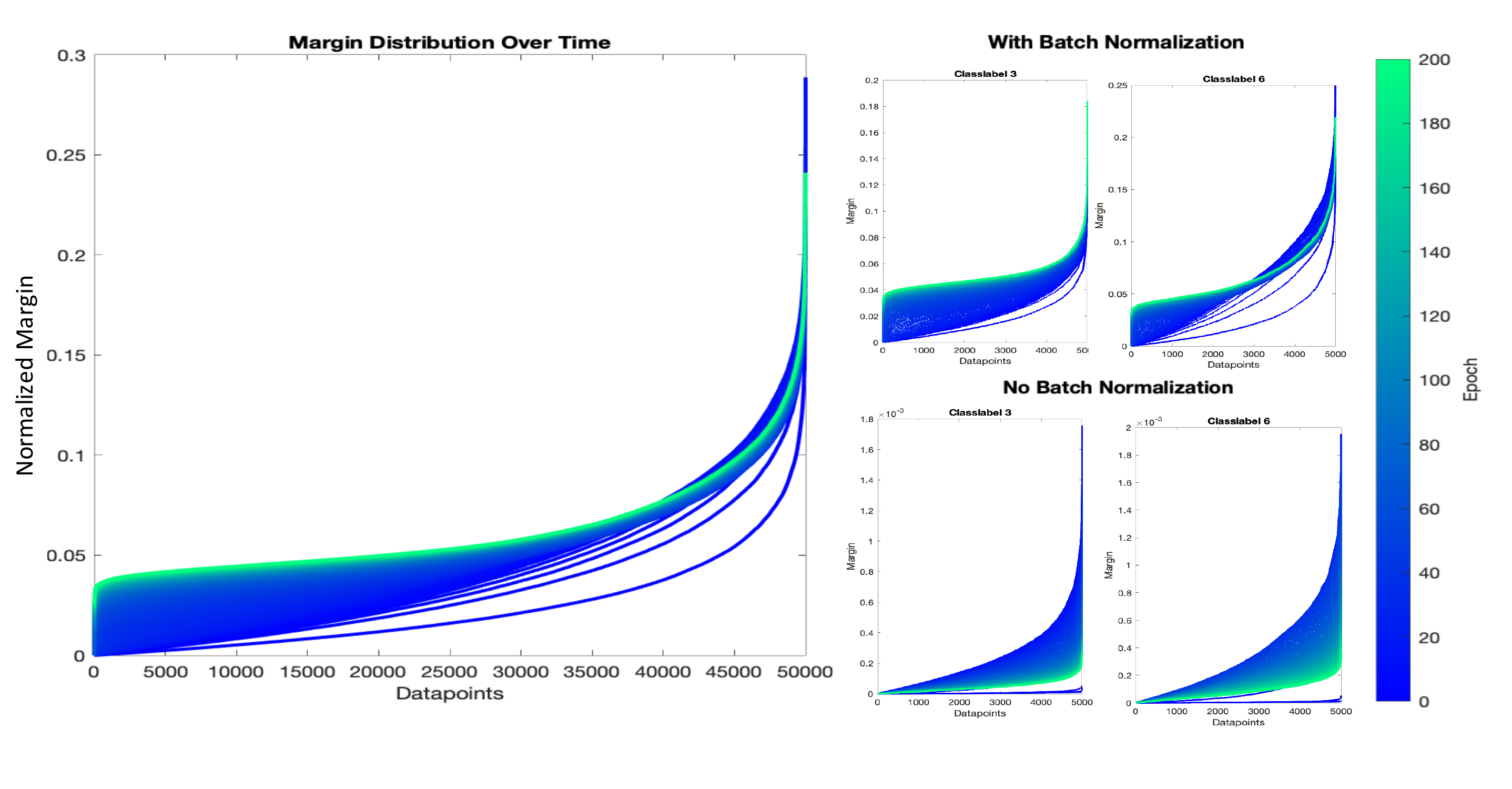} 
		\caption{\it \textbf{Margin Distribution over Time} 
		A convolutional network (using the architecture and parameters in Section \ref{expmethods}) was trained on CIFAR10, the margin distribution was recorded and sorted at every epoch. On the left, we see the margin distribution of all 50,000 datapoints from epoch 1 (dark-blue) to epoch 200 (light-green) as indicated by the color bar on the far-right. To explore the results in neural collapse, we visualize the margin distribution of individual classes for networks trained with batch normalization and without. Some class-labels seem to have more of an effect on the distribution for that label than others but this is dependent on batch normalization. 
		}
		\label{main3}
		\vspace{-0.45cm}
	\end{figure*}

	Do these results mean that we are finding no NC1? On the left
        of Figure \ref{main3}, we find that in the presence of both
        regularization and batch normalization, the distribution of
        margins does indeed get flatter with time. In further
        experiments we found however that without either L2
        regularization or batch normalization, such flattening is not
        apparent, see Sup. Mat. This is in line with all the
        experiments in  \cite{Papyan24652} using both of these
        techniques, and suggests that Neural Collapse relies on both
        regularization and normalization in agreement with the
        theoretical predictions of \cite{4625}.
	
	To further explore the relationship of margin distribution in the context of Neural Collapse, we visualize and analyze the margin distribution for individual classes. On the top-right of Figure \ref{main3}, we observe that for some classes, the margin distribution shifts and flattens more than for other classes. For class-label 9, the margin distribution shifts and flattens more over time (going from blue at epoch 0 to green at epoch 200) than for class-label 3. This effect is absent if we do not use batch normalization, as shown in the bottom right. Thus, although margin distributions of individual classes are potentially shifting the margin distribution at different scales, this is dependent on batch-normalization and regularization (see Sup. Mat. for regularization experiments), as suggested in \cite{4625}.

	\begin{figure*}[t!]\centering
		\includegraphics[trim = 0 0 0 0, width=1.0\textwidth, clip]{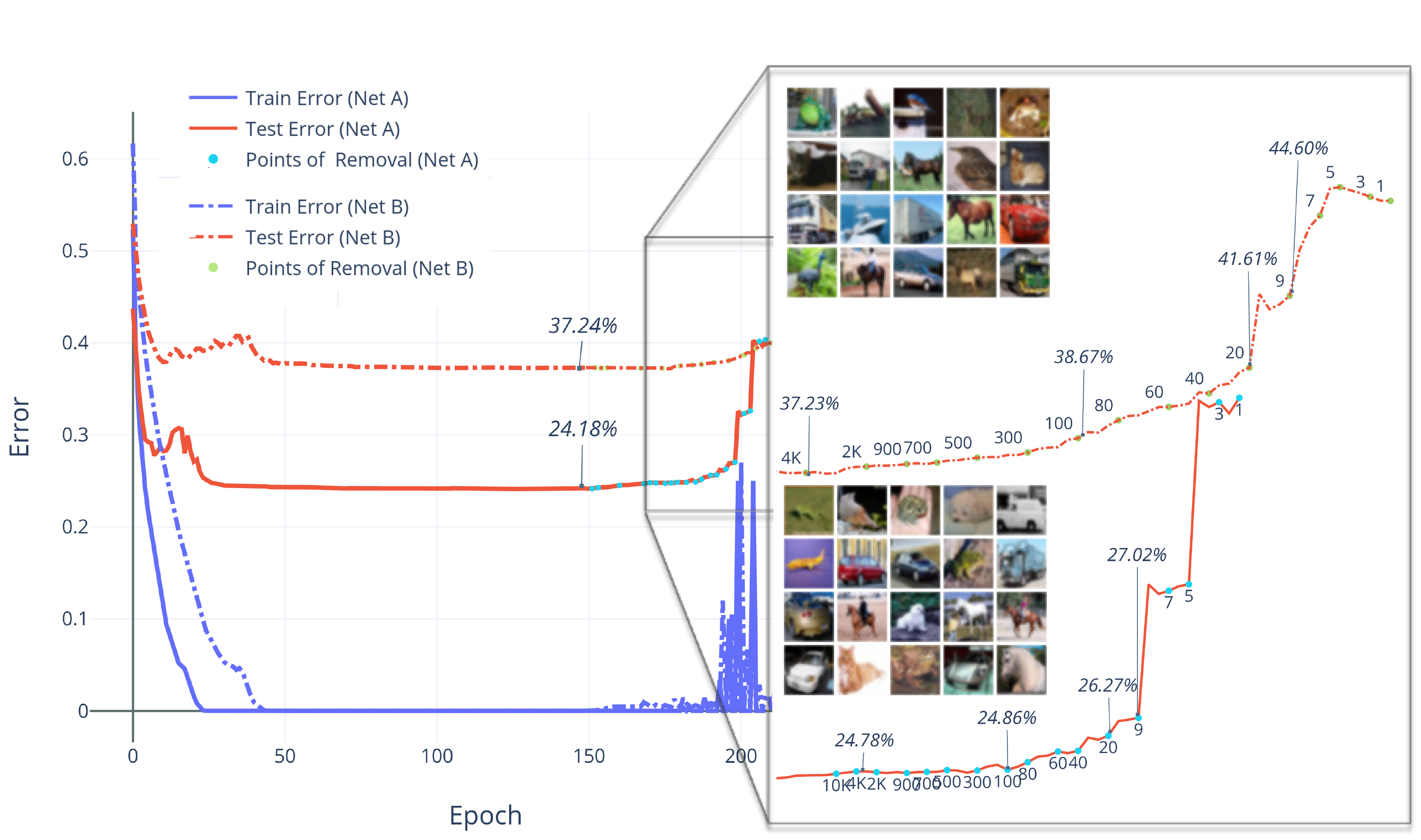} 
		\caption{\it \textbf{Data Points Removal Algorithm} Networks (A) and (B) were randomly initialized from a normal distribution with 
			$std$ of 0.01 and 0.09, respectively. This leads the networks to converge to different minima with different test errors ($\sim13\%$). 
			After data separation and full convergence, the algorithm ranks the datapoints based on their margin size and starts removing
			datapoints during further training, removing those with highest margins at every point. At every blue and green point, a set of points were 
			removed, with the numbers displaying the amount of datapoints left in the training set. The right side 
			zooms in to show that the test error does not significantly change until 100 datapoints are left (by 0.68\%  for (A) and 1.43 \% for (B)). The test
			error changes more when there are only 20 datapoints left for both (A) and (B). The figure inserts show these 20 datapoints for both networks. 
			The two sets are different, showing that the networks converged to infima with different support.
		}
		\label{main4}
	\end{figure*}

	\section{Compressing the training set dynamically}

	As suggested by the notion of stability \cite{kn2002}, datapoints close to the
	separation boundary are crucial for good test
	performance.  Data with large margin, however, do not
	contribute to stability.
	It has been long observed that training long past the time of
	achieving the separation of the data (i.e.
	$0\%$ training error) leads to improved test
	performance. This has been understood
	\cite{2017arXiv171010345S} to result from the fact that while
	the training classification accuracy converges fast, the
	margin converges much slower. As we keep training the network
	past separability, the margin keeps improving
	\cite{theory_III,2019arXiv190507325S,DBLP:journals/corr/abs-1906-05890}
	(see also the Sup. Mat.), with the largest
	contributions to this improvement coming from the datapoints
	with the current smallest margin.

	These theoretical considerations suggest that after we have
	separated the data, we should be able to safely drop training
	datapoints with large margin.  The question now is: how much
	can we compress, without spoiling the generalization
	performance? We can see in Figure \ref{main4} that gradually removing
	all but 200 datapoints with the smallest margin has no impact
	on the test performance for a well performing network (solid red line), if we start removing the data after
	separation has been achieved.  More interestingly, we get a very
	minimal drop in performance ($\sim2\%$) when we keep only 20
	datapoints in the training set. We find that minima that perform
	better can be compressed more. Figure \ref{main6} compares two
	minimizers with different test performance: we can readily see
	that further removal of datapoints leads to more degradation
	of performance for the network with worse test error,
	as compared to the better performing network.  We thus have a
	demonstration that, at finite times, larger number of small margin data leads to worse test performance.

\begin{figure}
		\includegraphics[trim = 0 0 30 0, width=1.0\columnwidth,clip,draft=false,]{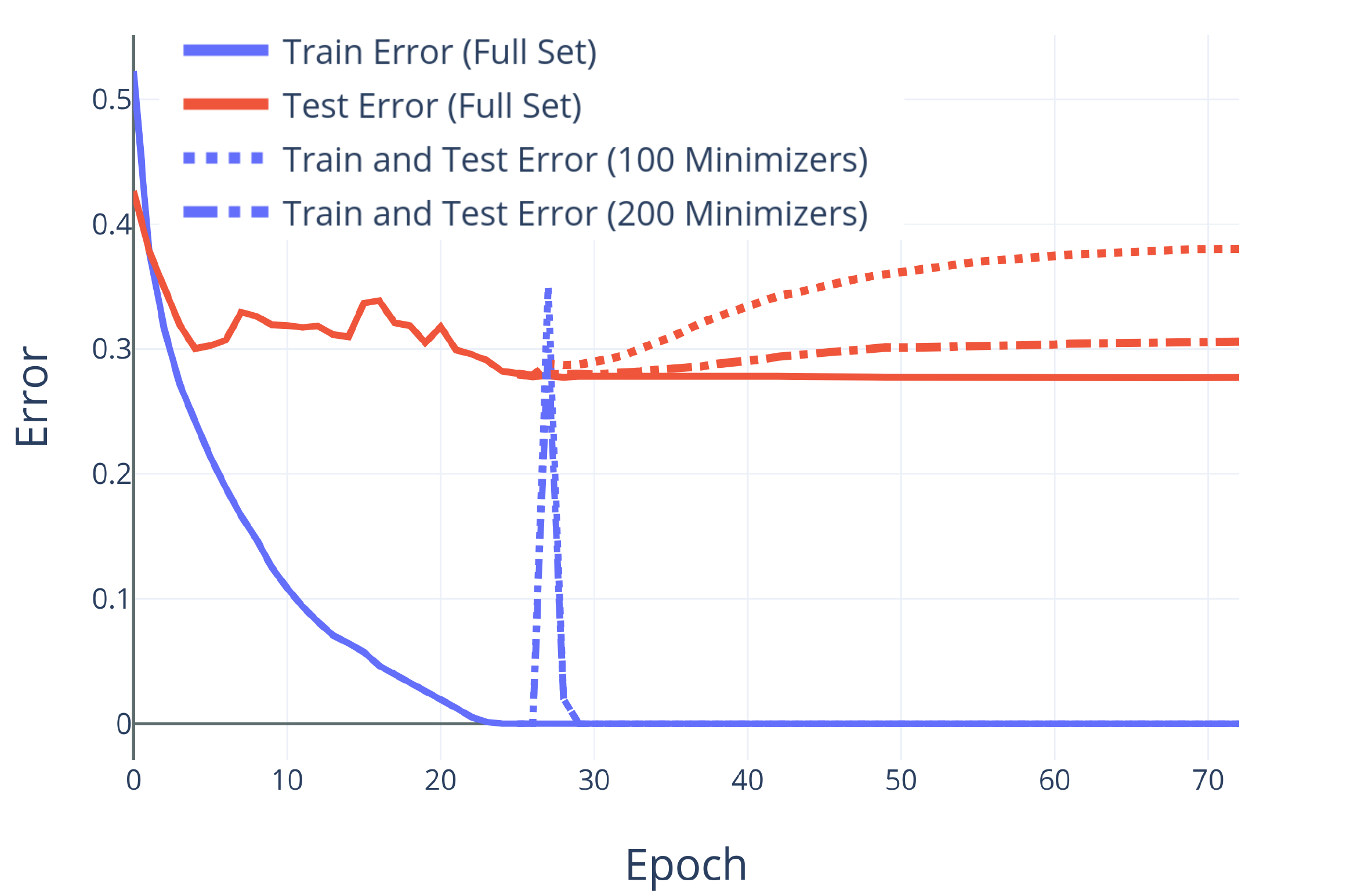}
		\caption{\it \textbf{Compression After Data Separation} 
			During training, right after data separation, datapoints with the large margins were removed, leaving either 100 or 200 datapoints with the smallest margins. 
			When the dataset is compressed to 200 datapoints the test error increases slightly but plateaus to a good test performance for the network architecture (2.89 \% change).
		}\label{main5}	
		\end{figure}

	This approach is indirectly related to data distillation that has been
	studied in \cite{DBLP:journals/corr/abs-1811-10959}. There,
	the authors noticed that it is possible to train a CIFAR10
	classifier on just 10 synthetic datapoints and achieve 54\%
	accuracy on the test set. Unlike in the case of data
	distillation however, here we first train the network to the
	point of separability and by gradually removing datapoints
	achieve {\it no significant drop in
		performance up to keeping $\sim200$ datapoints}. 
	These results strongly suggest a novel training scheme for speeding up convergence, in which we remove
	most of the training data (those with large margin) right after reaching separation. Rapid compression of the dataset down to 100 examples hurts performance, but keeping 200 points
	only leads to $\sim2.89\%$ reduction of accuracy for our CNN architecture trained with SGD, a batch-size of 50 and a learning rate of 0.01, see Figure \ref{main5}.
A more thorough search of the hyperparameter space reveals that with large learning rates and small batch size, compression of CIFAR10 down to 200 data
can lead to drop in performance as low as 0.18\% for our CNN and 0.11\% for DenseNet-BC, as seen in Table \ref{compression-table}. For more details on the experiments, see Sup. Mat.

\begin{table}[]
\caption{Drop (in \%) of test performance after compression.}
\begin{sc}
\resizebox{\columnwidth}{!}{
\begin{tabular}{|c|r|r|r|}
\hline
\multicolumn{4}{|c|}{\textbf{Convolutional Network (CIFAR10) with SGD}}\\ \hline
\multicolumn{1}{|c|}{\textbf{\diagbox{Batch Size }{Learning Rate}}} &\multicolumn{1}{c|}{$10^{-1}$} & \multicolumn{1}{c|}{$10^{-2}$} & \multicolumn{1}{c|}{$10^{-3}$} \\ \hline
200& 18.29& 25.15& X\\ \hline
100& 2.55& 22.18& 11.810\\ \hline
50& 1.68& 4.84& 11.55\\ \hline
20& 0.45& 4.21& 8.99\\ \hline
10& 1.75& 0.86& 6.48\\ \hline
1& 0.18& 0.46& 1.91\\ \hline
\end{tabular}
}
\end{sc}
\vskip -0.15in
\label{compression-table}
\end{table}		
	
	\begin{table}[]
\resizebox{\columnwidth}{!}{%
\begin{sc}
\begin{tabular}{|c|l|l|l|l|l|l|}
\hline
\multicolumn{7}{|c|}{\textbf{Dense Network (CIFAR10)}}\\ \hline
\multicolumn{1}{|c|}{\textbf{Optimizer}}&\multicolumn{3}{c|}{Adam}& \multicolumn{3}{c|}{SGD}\\ \hline
\multicolumn{1}{|c|}{\textbf{\diagbox{Batch Size}{Learning Rate}}} & $10^{-1}$ & $10^{-2}$ & $10^{-3}$ & $10^{-1}$ & $10^{-2}$ & $10^{-3}$ \\ \hline
200& 7.25& 2.78& 4.11& 1.54& 5.99& 33.31\\ \hline
100& 1.88& 0.95& 2.70& 0.72& 3.94& 3.60\\ \hline
50& 0.12& 0.85& 1.72& 0.11& 2.38& 4.08 \\ \hline
\end{tabular}
\end{sc}
}
\vskip -0.15in
\end{table}

	\subsection{Similar Initialization Leads to Different Important Data}
	
	\begin{figure}[t]
		\includegraphics[trim = 0 0 30 0, width=1.0\columnwidth,clip,draft=false,]{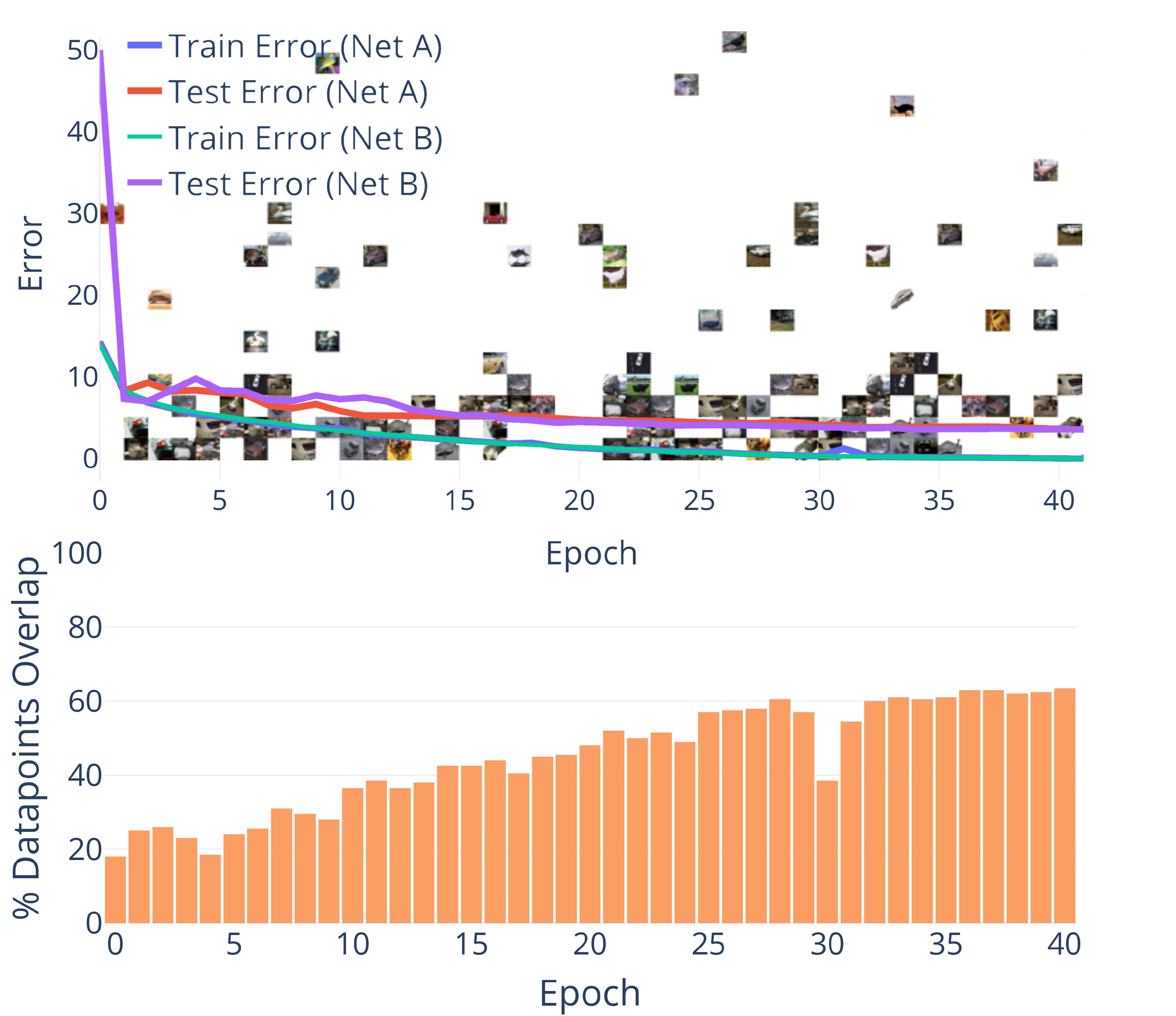}
		\caption{\it \textbf{Same Initialization - Different small margin datapoints} 
		Two networks were initialized from the same margin distribution (standard deviation 0.01) and trained on CIFAR2. Both networks converge to similar minima but the exact datapoints with the smallest margins differ. On the top, the errors are depicted and the vectorized images (cars and birds) represent the overlapped datapoints in the smallest 20 -- there are only a few overlapping points. The percentage of overlapping datapoints for the smallest 200 margins are depicted at the bottom (60 \% of them overlap).
		}\label{main6}	
		\vspace{-0.2in}
		\end{figure}

	It might be tempting to try to understand why certain datapoints seem to drive the dynamics in the terminal phase of training (post data separation). After all, we can see in Figure \ref{main4} that minima of different levels of test performance have a totally different set of small margin data. The converse does not seem to be however true, as can be seen in Figure \ref{main6}, where we initialized the network several times from the same statistical distribution (normal with std 0.01). We find that the initial randomness propagates through the training procedure, leaving us with similarly performing minima (3.54\% and 4.65\% test errors) with different smallest margin data. 
	
	On the top of Figure \ref{main6}, we can see the training and test error of two networks trained on CIFAR2. The vertically vectorized images at each epoch represent the overlapping set of the 20 datapoints with thee smallest margins ("support vectors") between the two networks. Visually, we can see there are not many overlapping datapoints. The bottom plot shows the percentage of overlapping datapoints in the 200 smallest margin datapoints of the networks. Although over time the percentage of overlapping data increases, only 63 percent of the same datapoints in this bin of 200 are present in both networks at margin convergence (last epoch). 
	
\begin{figure}
    \centering
\includegraphics[trim = 5 0 0 20, width=0.85\columnwidth, clip]{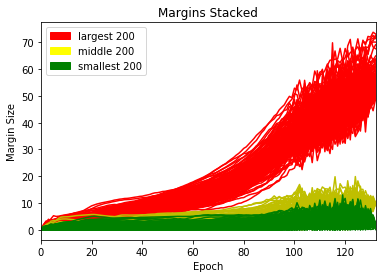}    
    \caption{	\it \textbf{Visualization of the 200 smallest, middle and largest margins over time}. Data separation occurs at epoch 40. This network was trained with a batch size of 1 using SGD with a learning rate of 0.01. There is no clear way of predicting which datapoints will have smallest margin before data separation.}
    \label{main7}
    \vspace{-0.2in}
\end{figure}

	Not only does the randomness at initialization play a role, but even more importantly, we find that for a given training run, we cannot reasonably predict which data are going to support the dynamics the most until just before data separation takes place, since only 40 percent of the support vectors at margin convergence are present at data separation, see Sup. Mat.  Figure \ref{main7} shows the margins of 600 datapoints (200 smallest, 200 in the middle and the largest ones) throughout training. We see from this that before data separation happens, it seems impossible to discover using only the margin information which datapoints will end up having the smallest margin.

	\section{Discussion and Conclusions}
	
	Recent theoretical results
        \cite{2019arXiv190507325S,DBLP:journals/corr/abs-1906-05890,PLB2020natcom}
        have shown that gradient descent techniques on
        exponential-type loss functions converge to solutions of
        locally maximum classification margin for overparametrized
        deep networks. In this paper we studied the distribution of
        margins on the entire training dataset and demonstrated that
        the area under the distribution is a good approximate measure
        for ranking different minima of the same network.
	
	Inspired by the recent theory\cite{4625} predicting various
        properties of Neural Collapse \cite{Papyan24652}, we
        investigated the prediction that none of the training data
        contribute more towards good generalization performance than
        others, at least asymptotically. We found that while on long
        timescales the distribution does get flatter, the dynamics
        effectively depends only on a few datapoints.  Once separation
        sets in, we can successfully compress most of the training
        datapoints (removing those $\{x_n, y_n\}$ with largest margins
        from the training set), going down from 50k examples to less
        than 200, without compromising on performance. In fact, since
        the property NC1 implies that we could in
        principle compress the training dataset down to one datapoint
        per class, very much in line with the distillation results of
        \cite{DBLP:journals/corr/abs-1811-10959}. Thus, in the
        presence of SGD and both batch normalization and L2 regularization
        (all three seem important for Neural Collapse to happen), we
        expect that all data are equally important to classification.
	
	In practice, we find that the compressed dataset we can obtain
        is highly dependent on the randomness of initialization. An
        obstacle to effectively predicting a good compressed set is
        the fact that the relevant datapoints only emerge around the
        time of data separation. This means that the algorithm for
        compressing the training dataset does not provide a massive
        speed boost, as one could hope. More importantly however,
        these results cast doubt on the endeavour of interpreting the
        features that a network learns -- we can expect that the
        randomness of which training points contribute the most to the
        solution of the optimization problem implies that the "high
        capacity" features most relevant to classification are also
        random and inconsistent between different training runs.
	
	The results in this article motivate potential more
        fine-grained investigations into the early pre-separation
        dynamics -- if we could earlier predict the compressed
        dataset, we would have a way to drastically speed up
        training. Additionally, it would be interesting to understand
        why dataset compression is so successful with small batch size
        and large learning rate -- is it connected to the suggestion
        in \cite{PoggioCooper2020} that small batch SGD is more likely
        to find global minima of the loss?

	{\bf Acknowledgments} This material is based upon work supported by the Center for
	Minds, Brains and Machines (CBMM), funded by NSF STC award
	CCF-1231216.

	\newpage
%

	
	%
	%
	%

\bibliography{Boolean}
\bibliographystyle{icml2021}

\appendix

	\section{Empirical Risk Minimization}

	We recall a few basic  definitions from  \cite{Mukherjee2006}
	about Empirical Risk Minimization as a class of algorithms for
	supervised learning. 
	
	We assume there exists  an unknown probability distribution $\mu(x,y)$ on the
	product space $Z = X \times Y$.  We assume $X$ to be a compact
	domain in Euclidean space and $Y$ to be a closed subset of $\R^k$.
	The measure $\mu$ defines an unknown {\it true function} $T(x) =
	\int_Y y d \mu(y | x)$ mapping $X$ into $Y$, with $\mu(y | x)$ the
	conditional probability measure on $Y$.

	We are given a training set $S$ consisting of $n$ samples (thus $|S|
	= n$) drawn i.i.d.  from the probability distribution on
	$Z^n$, with $ 	S = (x_i,y_i)_{i=1}^n = (z_i)_{i=1}^n. $ 
	
	The basic goal of supervised learning is to use the training
	set $S$ to ``learn'' a function $f_S$ that evaluates at a new
	value $x_{new}$ and (hopefully) predicts the associated value
	of $y_{pred} = f_S(x_{new}). $ In this paper we consider the binary pattern classification
	case in which $y$ takes values from
	$\{-1,1\}$e.

	In order to measure goodness of our function, we need a loss function
	$\ell(f,z)$.
	
	Given a function $f$, a loss function $\ell$, and a probability
	distribution $\mu$ over $X$, we define the {\it expected error}
	of $f$ as:
	$$ 
	I[f] = \E_z \ell(f,z)
	$$ 
	
	In the following we denote by ${S^i}$ the training set with the point
	$z_i$ removed and ${S_{i,z}}$ the training set with the point $z_i$
	replaced with $z$.  For Empirical Risk Minimization, the functions
	$f_S$, $f_{S^i}$, and $f_{S_{i,z}}$ are almost minimizers (see
	Definition \ref{almostERM}) of $I_{S}[f]$, $I_{S^i}[f]$, and
	$I_{S_{i,z}}[f]$ respectively.

	In the following, we will use the notation $\pr_{S}$ and $\E_{S}$ to
	denote respectively the probability and the expectation with respect
	to a random draw of the training set $S$ of size $|S| = n$, drawn
	i.i.d from the probability distribution on $Z^n$. 
	
	Given a function $f$ and a training set $S$ consisting of $n$ data
	points, we can measure the {\it empirical error (or risk) of $f$}
	as:
	$$
	I_S[f] = \frac{1}{n} \sum_{i=1}^n \ell(f,z_i). 
	$$

	\begin{defn} \label{almostERM} 
		Given a training set $S$ and a function space $\cal{H}$, we define
		almost-ERM (Empirical Risk Minimization) to be a {\it symmetric}
		procedure\footnote{\begin{defn} \label{symm} 
				An algorithm is defined as symmetric if over training sets $S$
				$$\E_S \ell(f_S,z) = \E_{S,\pi} \ell(f_{S(\pi)},z),$$ 
				for any $z$ and $S(\pi) = \{z_{\pi(1)},...,z_{\pi(n)}\}$ for
				every permutation $\pi$ from $\{1,...,n\}$ onto itself.
		\end{defn}} that selects a function $f^{\varepsilon^E}_{S}$ that {\it almost
			minimizes} the empirical risk over all functions $f \in \cal{H}$,
		that is for any given $\varepsilon^{E} > 0$:
		\begin{equation}
		\label{erm}
		I_S[f^{\varepsilon^E}_S] \leq  \inf_{f \in \cal H} I_S[f] + \varepsilon^E. 
		\end{equation}
		
	\end{defn}

	In the following, we will drop the dependence on $\varepsilon^E$ in
	$f^{\varepsilon^E}_S$.  Notice that the term ``Empirical Risk
	Minimization'' (see Vapnik \cite{Vapnik98}) is somewhat misleading:
	in general the minimum need not exist. In fact, it is precisely for this reason that we use the notion of almost minimize given in equation (\ref{erm})  since the infimum of the
	empirical risk always exists. 
	
	We will use the following notation for the {\it loss class} $\cal L$
	of functions induced by $\ell$ and $\cal H$. For every $f \in {\cal H}$,
	let $\ell(z) = \ell(f,z)$, where $z$ corresponds to $x,y$. Thus
	$\ell(z): X \times Y \rightarrow \R$ and we define ${\cal L} =
	\{\ell(f): f \in {\cal H}, L \}$.

	Remark:{\it
		In the learning problem, {\it uniqueness} of the
		solution of ERM is always meant in terms of uniqueness of $\ell$
		and therefore uniqueness of the equivalence class induced in $\cal
		H$ by the loss function $L$. In other words, multiple $f \in {\cal
			H}$ may provide the same $\ell$. Even in this sense, ERM on a
		uGC class is not guaranteed to provide a unique ``almost
		minimizer''. Uniqueness of an almost minimizer therefore is a
		rather weak concept since uniqueness is valid {\it modulo the equivalence
			classes} induced by the loss function {\it and} by
		$\varepsilon$- minimization.}
	
	\subsection{Ramp loss}
	In the main part of the paper, we use the ramp loss, defined in  \cite{2017arXiv170608498B} as 
	\begin{equation*}
\ell_{\gamma}(y, y')= 
\begin{cases}
1, & \text{if} \quad yy' \leq 0, \\
1-\frac{yy'}{\gamma}, & \text{if} \quad 0 \leq yy' \leq \gamma, \\
0, & \text{if} \quad yy' \geq \gamma.
\end{cases}
\end{equation*}

We define $\ell_{\gamma=0}(y, y')$ as the standard $0-1$ classification
error and observe that $\ell_{\gamma=0}(y, y') <\ell_{\gamma>0}(y, y')$.

	\section{Invariance of data-fitting under deletion of datapoints}
	In the paper we study stability by removing datapoints from the training set.  One obvious question is whether the critical points of the network trained on the whole dataset are also critical points
	for the network trained on the smaller set. The answer is to this is affirmative for the solutions that fit the data, as can be seen below.
	
	Let us start with the simpler case of regression, i.e. the square loss $L = 1/N \sum_n \left(f(W;x_n)-y_n\right)^2$. The interpolating solution
	$W^*$ is the global minimum satisfying $f(W^*;x_n)=y_n \ \forall n$. It is straightforward to see that if $f(W^*;\cdot)$ fits $\{\{x_1,y_1\},\ldots,\{x_n,y_n\}\}$,
	it will also fit all the data sans $\{x_i,y_i\}$ for any $i$. For non-interpolating minima, however, we need to satisfy
	\begin{equation}
	\sum_n \left(f(W;x_n)-y_n\right) \nabla_w f(W,x_n) = 0,
	\end{equation}
	with $E_n\equiv \left(f(W;x_n)-y_n\right) \neq 0$. Notice that this leaves us two possible situations -- either $\nabla_W f(W;x_n) = 0$ or the linear combination 
	vanishes. In the first case, for ReLU networks, it follows by the structural property  that $ f(W;x_n) = 0$, which is a trivial solution. The linear combination
	on the other hand is immediately seen to be unstable to removing a datapoint -- you cannot remove a single nonzero term from a vanishing sum and have it still be zero.
	Hence, in the case of regression, interpolating minima are invariant to removing datapoints from the training set.
	
	In the case of classification, the story is a bit more involved, since strictly speaking global minima are at infinity. We can consider however the
	dynamics of normalized weights $V$, in which case we have the condition (for exponential loss)
	\begin{equation}
	\sum_n e^{-\rho y_nf(V;x_n)} y_n\left(\frac {\partial f(V;x_n)}
	{\partial V_k} - V_kf(V;x_n)\right)=0.
	\end{equation}
	For $\rho$ large enough, the main contributions to this sum come from the datapoints for which $y_nf_V(x_n)$ is the smallest positive value. We see immediately that removing one of the other points will have a miniscule influence, but removing a datapoint with the smallest margin could lead
	to large changes, as none of the exponential coefficients $e^{-\rho y_nf_V(x_n)}$ are zero. 
	Thus removing datapoints for which  $V_k f_V(x_n) \neq \frac {\partial f_V (x_n)}{\partial V_k}$ does not leave us with the same critical point.
	What does, however, get preserved in this case is the notion of separability -- if $y_nf(V^*;x_n) > 0 \ \forall n$, then the smaller dataset is also immediately
	separable at the point $V^*$, even if that point is not necessarily a critical point.

	\section{Further experiments}	
	We provide here further experiments exploring the relation between margin maximization, the margin distribution and generalization performance.

	\subsection{Landscape of Smallest-Margin Datapoints}
	
	\begin{figure*}[ht!]\centering
		\includegraphics[trim = 40 10 100 100, width=1.0\textwidth, clip]{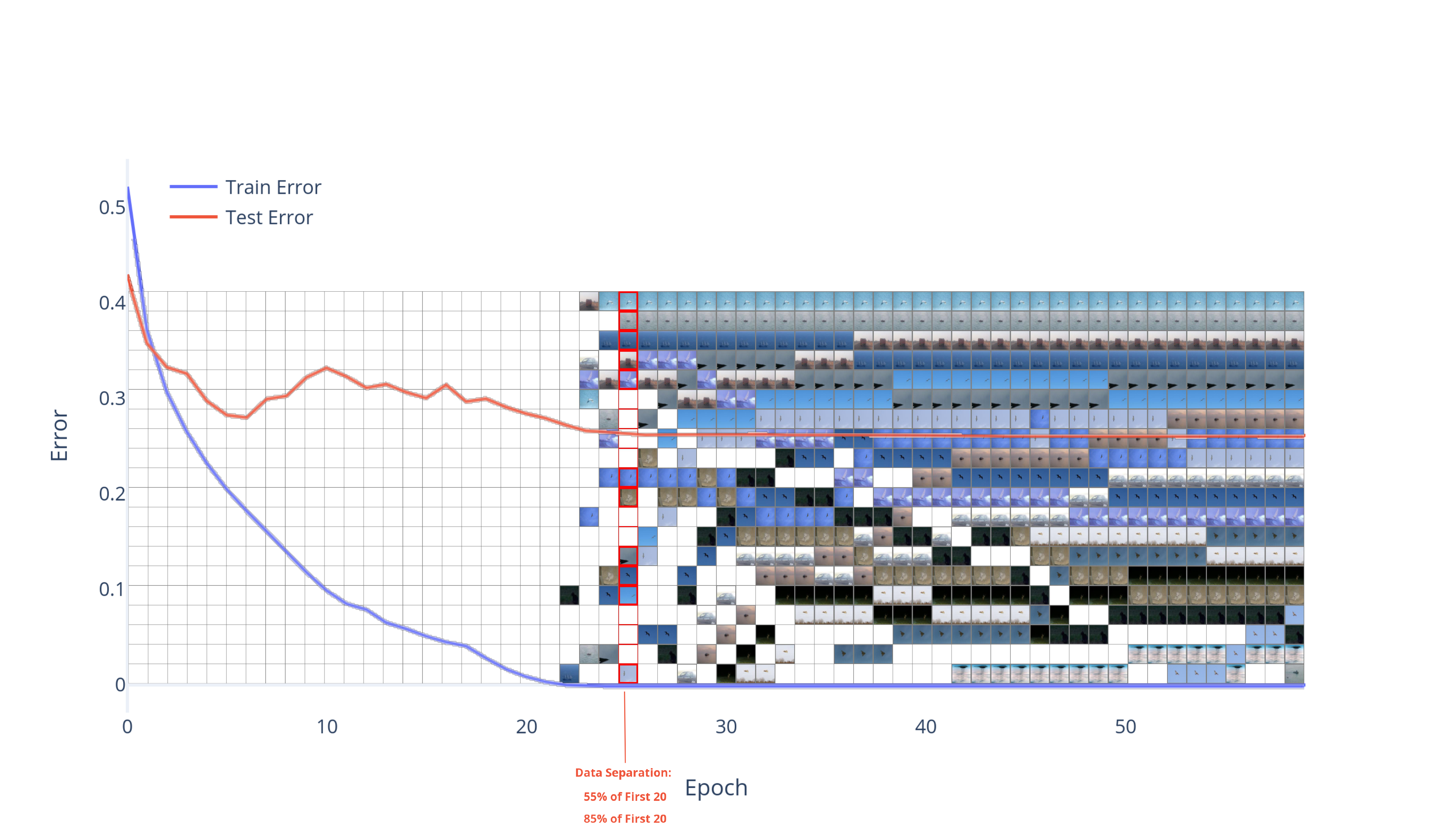} 
		\caption{\it \textbf{Datapoints with Smallest Margin (20)} A 6-layer neural network implemented in PyTorch was trained on the full CIFAR-10 dataset with Stochastic Gradient Descent
			(SGD) on cross-entropy loss. The figure shows the test and training error as well as the 20 datapoints with the smallest margin if those datapoints are in the set of datapoints in the 
			last epoch (set of datapoints the network converges to), otherwise the datapoints are not displayed. 
			At data separation, 55 percent of the first 20 datapoints in the last epoch are already present, suggesting that data compression 
			can be performed right after data separation. 
		}
		\label{main20}
	\end{figure*}
	
	In Figure 3 of the main text, we presented an algorithm that allows the removal of datapoints with large margins after data-separation and convergence with little effect on the test performance. 
	This algorithm requires the network to converge to a set a datapoints with the smallest margins, which usually happens much after data-separation but it results in a very small test performance decrease ($0.6\%$). 
	Here, we present experiments that guided the results in Fig 4 of main text, 
	where we showed that in the epoch right after data separation $99.6\%$ of the datapoints can be removed with a test performance decrease of only $2.9\%$. 
	
\begin{figure*}[ht!]\centering
		\hspace*{-0.8cm}
		\includegraphics[trim = 25 0 50 0, width=1.07\textwidth, clip]{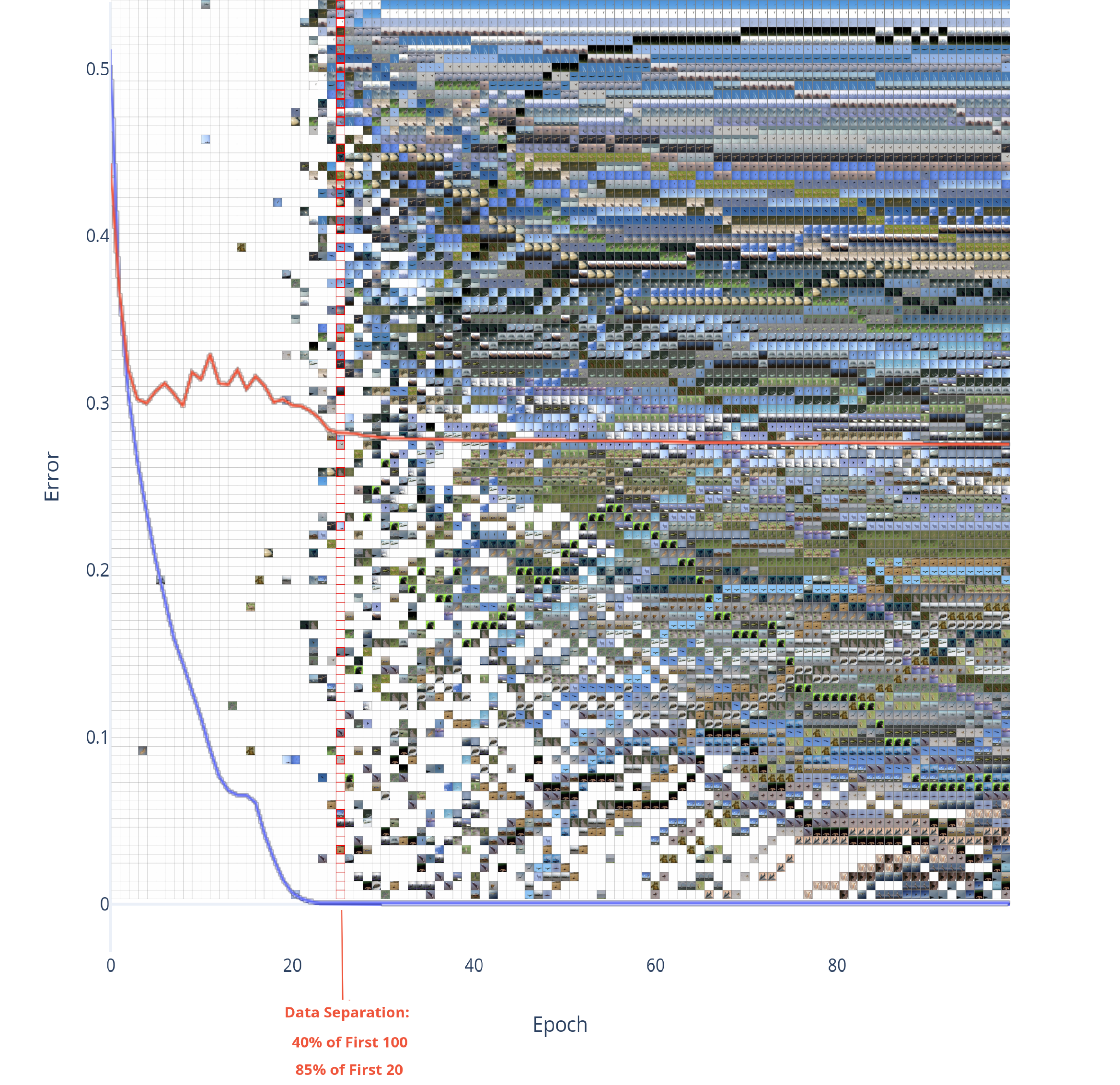} 
		\caption{\it \textbf{Datapoints with Smallest Margin (100)} 
			A 6-layer neural network implemented in PyTorch was trained on the full CIFAR-10 dataset with Stochastic Gradient Descent
			on cross-entropy loss. The figure shows the test and training error as well as the 100 datapoints with the smallest margin if those datapoints are in the set of datapoints in the 
			last epoch, otherwise the datapoints are not displayed. At data separation, 40 percent of the first 100 datapoints and 85 percent of the 20 datapoints in the last epoch are already present, suggesting that data compression 
			can be performed right after data separation. 
		}
		\label{main21}
	\end{figure*}

	The question we ask in the following experiments is what percentage of the set of smallest-margin datapoints to which the network converges to can be predicted right after 
	data separation. We take networks with the same architecture as those in the main text, trained on the full CIFAR10 dataset with SGD on cross-entropy loss 
	and extract the datapoints with the smallest margins at every epoch. We take the converged set of datapoints (last epoch on each figure) and check how many of them were present at each epoch and display those datapoints on Fig \ref{main20} and Fig \ref{main21}. The actual images are shown, the datapoints are ordered from the smallest margin starting at 
	the top to the largest at the bottom for every preceding epoch. The empty squares indicate that none of the datapoints from the set of converged small margin datapoints is present. The ordering of the datapoints
	continues to change after data-separation but a large number of them are present at data-separation. In Fig \ref{main20}, data-separation occurs at epoch 25. After taking the twenty datapoints with
	the smallest margin to which the network initially converges to (at epoch 60), we observe that $55\% (11/20)$ of those datapoints are present at the epoch of data separation and the first
	four datapoints change order only slightly throughout. Fig \ref{main21} shows a network initialized slightly different (through PyTorch's random initialization), where we 
	train the network for longer and instead of only extracting 20 of the datapoints with the smallest margins, we extract 100. Here, we take the set of datapoints to which the network converged to at epoch 100 and indicate how many of these are present at every preceding epoch. At data-separation, $85\%$ of the 20 datapoints with smallest margins and $40\%$ of the first 100 are present.
	
	These results suggest that more experiments like these could provide bounds for how early and how many datapoints can be removed during training without significantly affecting test performance. 
	Interestingly, the datapoints with the smallest margins for both experiments mostly include classes that are visually similar to the human eye (due to the backgrounds and CIFAR10 resolution), such as airplanes, birds and boats. 
	Further experiments can also give more insights on the dataset-dependence of the margin distribution.

	\begin{figure*}[ht!]\centering
		\includegraphics[trim = 0 0 0 0, width=1\textwidth, clip]{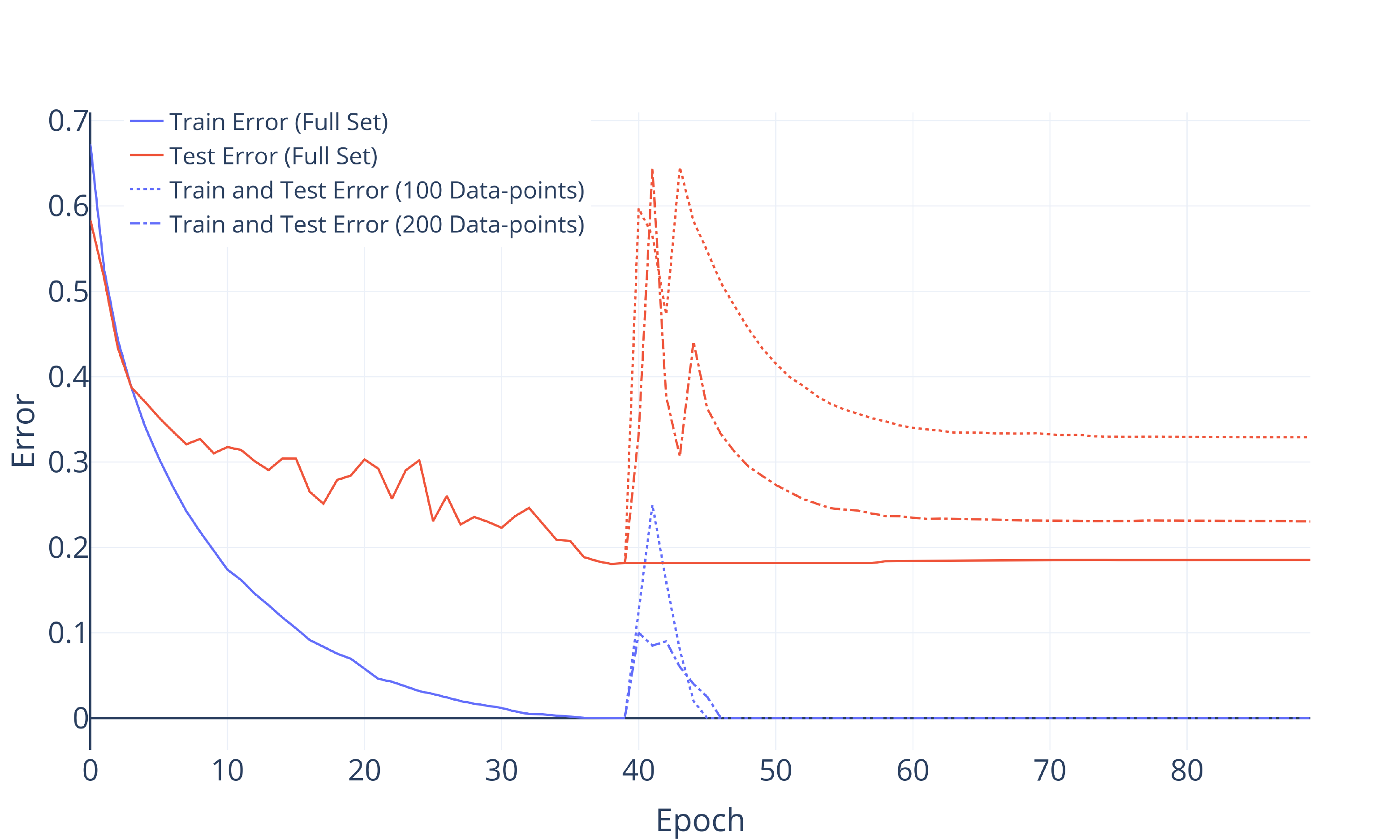} 
		\caption{\it \textbf{Compression After Data Separation (DenseNet)} 
			During the training of a DenseNet, right after data separation, datapoints with the large margins were removed, 
			leaving either 100 or 200 datapoints with the smallest margins. 
			When the dataset is compressed to 200 datapoints the test error goes slightly higher but plateus to a test performance accuracy only $4.49\%$ lower than 
			the network trained on the full dataset on the same number of epochs.
		}
		\label{main22}
	\end{figure*}
	
	Since many of the datapoints with smallest margins at convergence are present at data-separation, in Fig 4 of main text, we removed 49,9800 of the datapoints 
	and observed a decrease of test performance of $2.9\%$. To explore whether the same holds for architectures with higher test performance , we ran the same experiment
	but with a DenseNet that performs with around $90.2\%$ accuracy with data-augmentation and other optimizers. In this experiment, we only use SGD and perform
	no data-augmentation, which results in lower test performance ($82.7\%$) but still higher than the simple convolutional network presented in the main text. 
	Fig \ref{main22} shows that removing down to 200 datapoints with the smallest margins results in a decrease of test performance of $4.49\%$. This is a higher drop
	than for the convolutional network, but still results in higher performance than the network used in the main text. 
	In the future, we plan to explore this algorithm with different architectures and batch-sizes. Currently, after the removal of 
	datapoints, we use GD but it would be interesting to explore the test error with SGD on different batch sizes.

	\subsection{Replace-one stability experiment}
	\begin{figure*}[ht!]\centering
		\includegraphics[trim = 0 0 0 0, width=1.0\textwidth, clip]{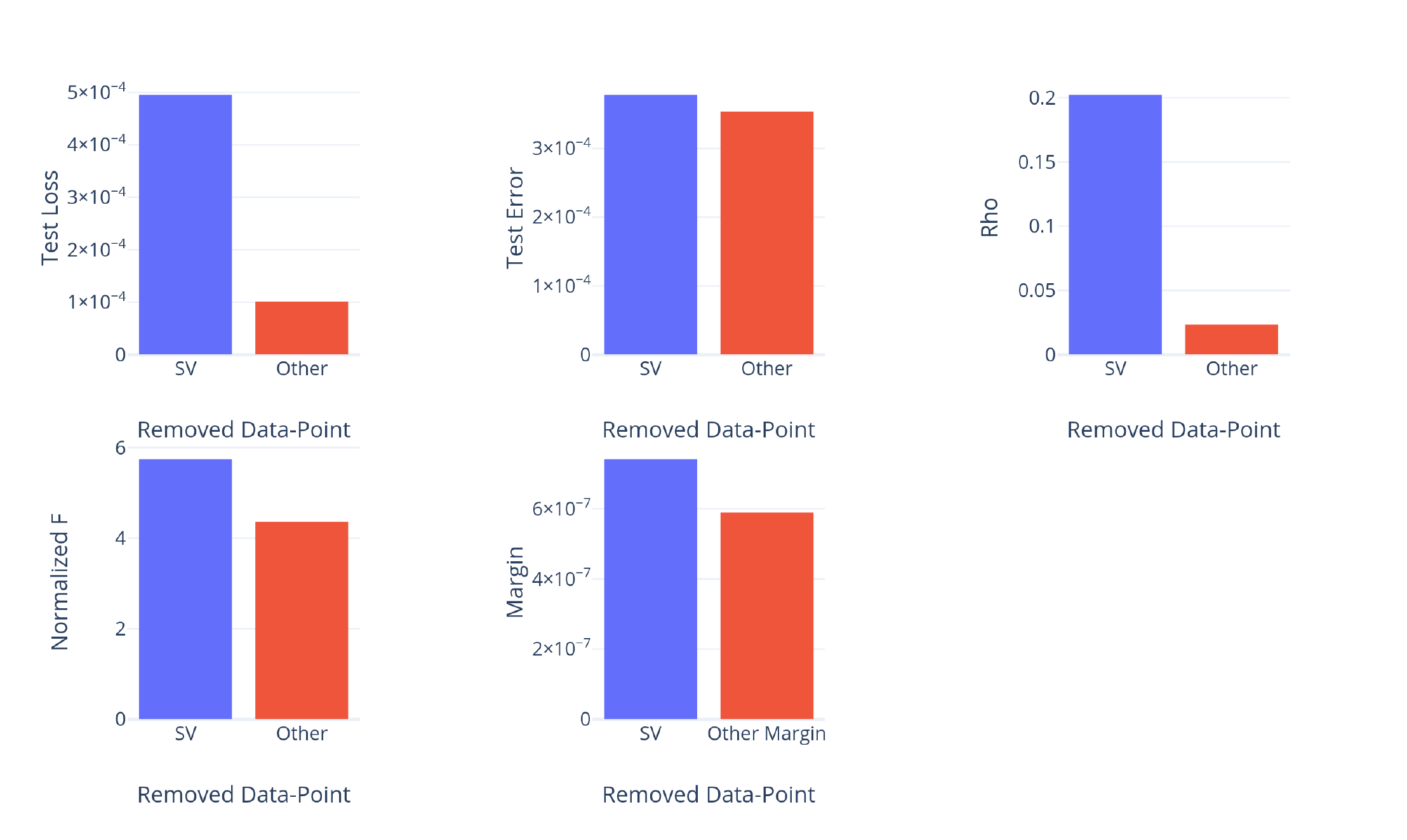} 
		\caption{\it \textbf{Replace-one Stability} We trained a network to full convergence and use the replace-one algorithm to 
			investigate the average difference caused by replacing the smallest-margin datapoint vs. replacing a random datapoint. The results show the
			average difference from the same network being trained continously on the full data-set vs. trained on $S^i$, the dataset 
			with one of the data points randomly replaced by one in the test set (and removed from the test set). Removing the smallest-margin datapoint 
			has a significant higher influence on the test loss and $\rho$.
		}
		\label{main23}
	\end{figure*}

	Here, we empirically explore the stability of a network with respect to the input data. We take a network trained with the full dataset and after data-separation 
	we replace one of the datapoints from the training set with one from the testing set and remove this datapoint from the testing set. We are interested in the differences 
	caused by replacing the datapoint with the smallest margin or any other random datapoint. We repeated this experiment on 1000 trials and obtained
	an average difference in test loss, test error, norm $\rho$ of the network, normalized output and the margin of the network for both the replacement of the 
	smallest-margin datapoint and a random datapoint. Fig \ref{main23} shows that removing the former results in a significantly higher difference on the test loss and $\rho$
	than replacing the latter.

	\subsection{True vs. random labels}
	In the main part of the article we discussed the distributions of margins for the network trained both on natural labels, as well
	as on randomized labels.  As observed in \cite{DBLP:journals/corr/ZhangBHRV16}, the network trained on random labels still has sufficient capacity 
	to separate the data fully, but it takes longer to converge to 0\% training error.
	
	This slower convergence to data separation can be also seen in the plot of the smallest margin for the two networks in Figure \ref{margin_conv}. 
	As observed in the main part of the article, the margin of the normalized network $f(V;x)$ trained on randomly labeled data is much smaller than the true labels case.
	It's interesting to note that the two networks can reach similar value of loss, but in the case of true labels this is done by maximizing the separability of data, and hence margin,
	while in the randomly labeled case the only way to reach small cross-entropy loss is through increasing the norm $\rho$, see \cite{liao2018surprising}.

	\begin{figure*}[ht!]\centering
		\includegraphics[trim = 0 0 0 0, width=0.8\textwidth, clip]{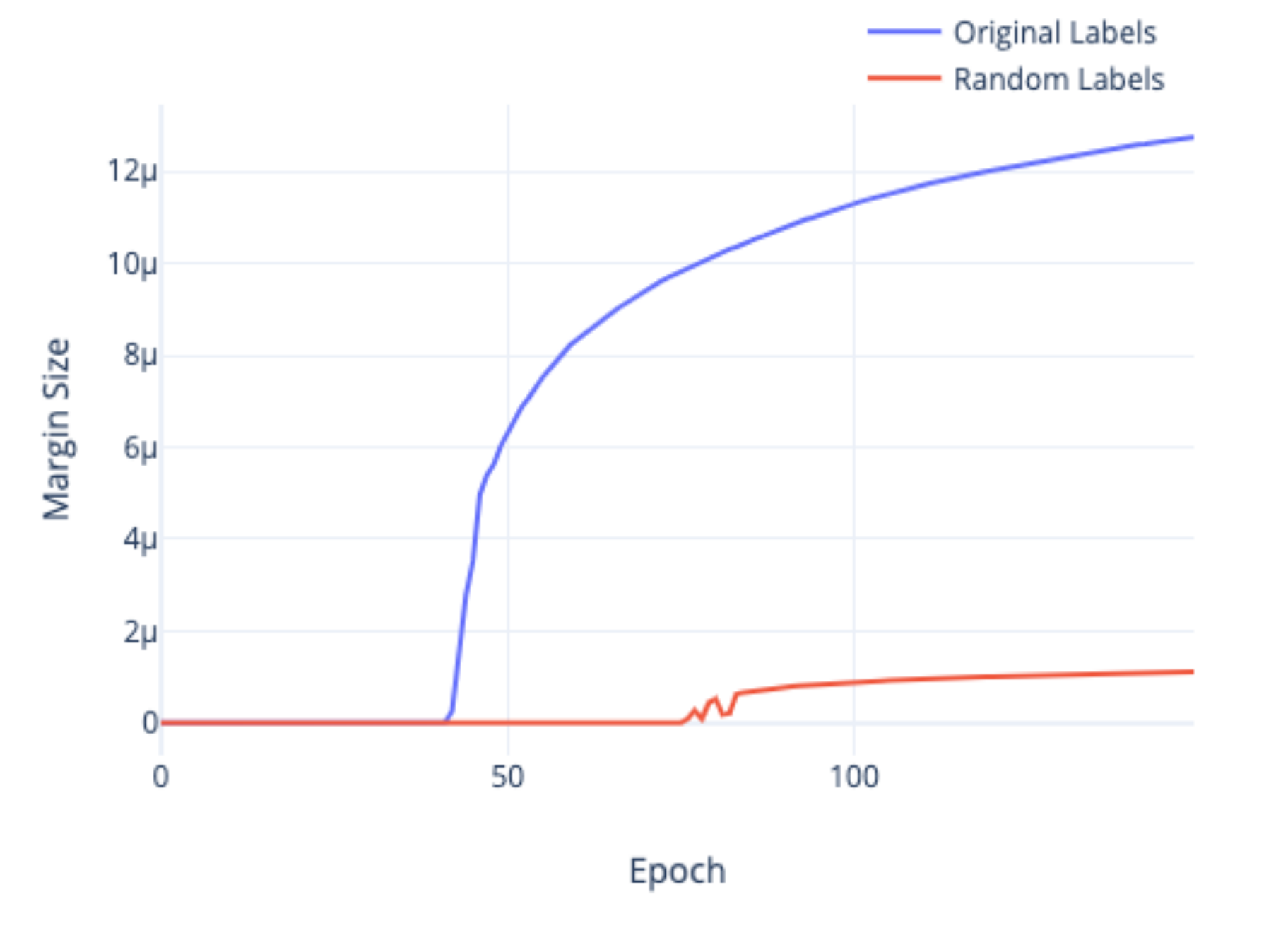} 
		\caption{\it \textbf{Natural and Random Labels - margin} Two 6-layer neural networks implemented in PyTorch were trained on the full CIFAR-10 dataset with Gradient Descent
			(GD) on cross-entropy loss, one with natural labels and the other with randomized labels. The figure shows the margin 
			$\operatorname*{argmin}( f_{y_{i}} - \max\limits_{j \neq i}f_{y_{j}})$ of the network, after data separation and full convergence. 
		}
		\label{supp2}
	\end{figure*}

	\subsection{Neural Collapse - Margin Distribution for Each Class}

\begin{figure*}[ht!]\centering
		\includegraphics[trim = 0 0 0 0, width=1.0\textwidth, clip]{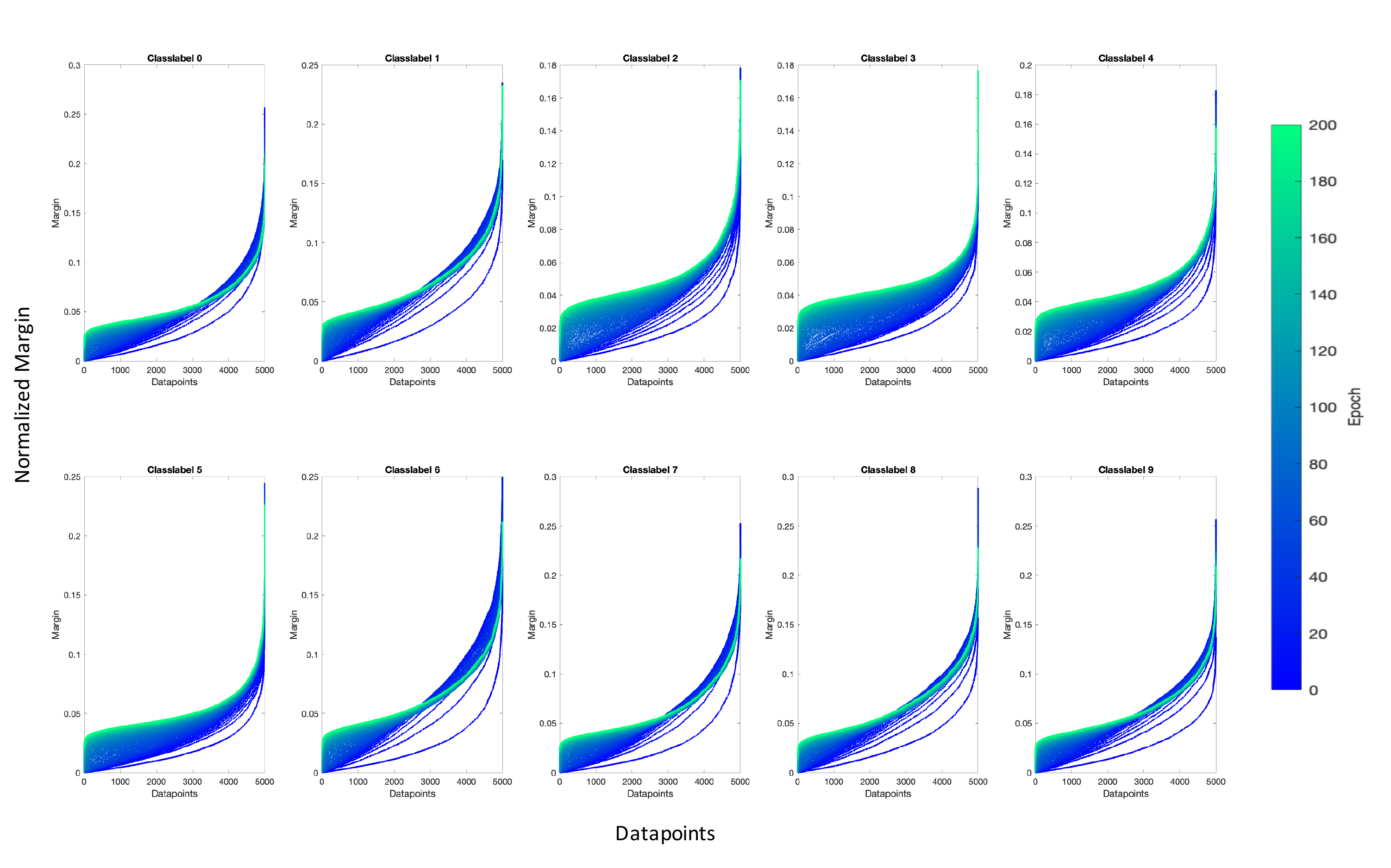} 
		\caption{\it \textbf{Margins for All Classes - With Batch Normalization} The network architecture described in the main text was trained with batch normalization until data separation and margin convergence (for 200 epochs - blue to green). The margin distribution over time is shown for each class label. For some classes, the margin distribution seems to shift and flatten over time but not for all. This shows evidence that specific datapoints are important for the overall distribution of a class, although which determining which points are important is not possible. This effect is increased with regularization. 
		}
		\label{classesnobatch}
	\end{figure*}

\begin{figure*}[ht!]\centering
		\includegraphics[trim = 0 0 0 0, width=1.0\textwidth, clip]{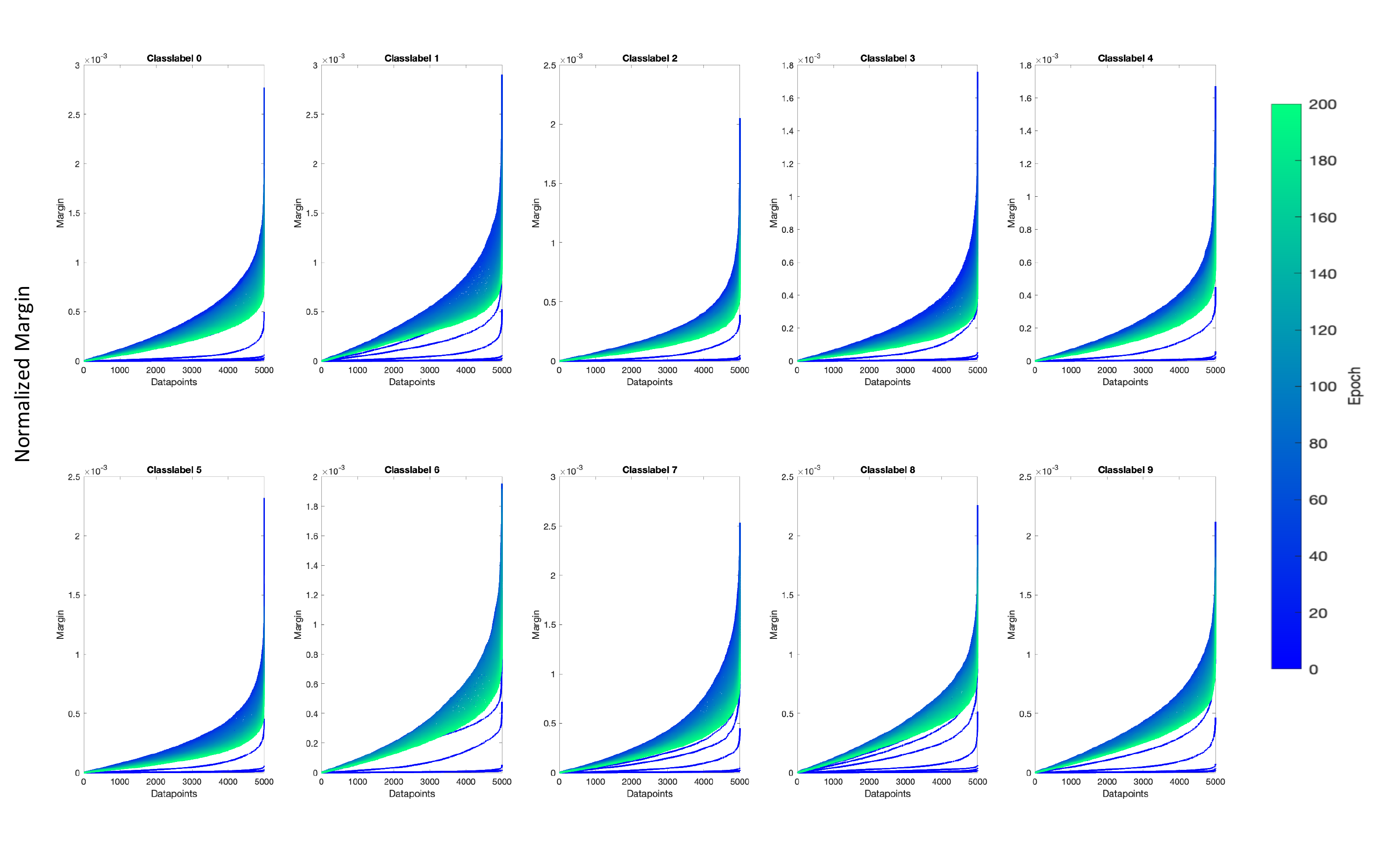} 
		\caption{\it \textbf{Margins for All Classes - With No Batch Normalization} The network architecture described in the main text was trained without batch normalization until data separation and margin convergence (for 200 epochs - blue to green). Unlike the network trained with batch normalization, the margin distribution does not seem to flatten.
		}
		\label{classesbatch}
	\end{figure*}

In the main text, we showed the change of the margin distribution over time. When networks are trained with batch normalization and regularization, the margin distribution shifts and flattens as the network converges. This is the case for some classes more than others. In the main text, we showed class 6 and 9 as examples. Here we show all classes for both a network trained with batch normalization (Fig \ref{classesbatch}) and without batch normalization (Fig \ref{classesnobatch}). 

There are several potential future questions these results raise with respect to Neural Collapse. A question we are exploring is why batch normalization is required for Neural Collapse and the flattening of the margin distribution over time. Another question to explore is why this behavior is true for some classes more than others; Is class-specific behavior dependent on initialization and hyper-parameters, or is it independent?

\subsection{More Details on Compression Experiments}

The continuous downsizing experiments implemented a continuous dataset downsizing regime. From the original 50,000 datapoints, we started by removing the 5,000 training samples with the largest margin until only 10,000 datapoints remained in the dataset. Then, we removed 1,000 at a time until 1,000 datapoints remained, followed by removing 100 at a time then by 10, and finally 1 at a time. After removing each chunk of datapoints, the network was retrained until reaching perfect separation again.

The immediate downsizing experiments implemented a one-time removal of datapoints. A model was trained to 100\% training accuracy on the original 50,000 datapoints. After convergence, the margins of the data points were calculated, and all except the 200 with the smallest margin size were removed.

Both sets of experiments were run with the CIFAR-10 dataset using PyTorch. We used a 5-layer CNN with 4 convolutional layers and one fully connected layer, with each of the former followed with a batch normalization layer and a ReLU nonlinearity layer. 

In the continuous removal experiments, along with standard SGD and the Adam optimizer, we used a learning rate of 0.01 and batch sizes of 254 until datapoint removal, at which point we performed full gradient descent. No data augmentation was performed for these networks. We additionally experimented with a DenseNet-BC implementation with 6, 12, 12, and 16 blocks. 

 In the immediate downsizing experiments, we used learning rates of 0.1, 0.01, and 0.001 with SGD, and batch sizes of 1, 10, 20, 50, 100, and 200. 

\subsection{Visualization of Margin Distributions}
\begin{figure}
    \centering
\includegraphics[width=0.5\textwidth]{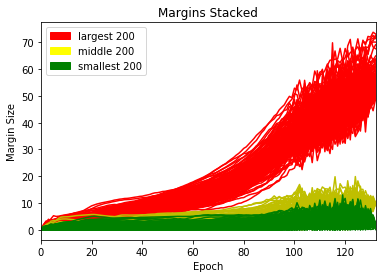}    
    \caption{Visualization of the datapoint margins during training. Data separation occurs at epoch 60. This network was trained with a batch size of 1 using standard SGD with a learning rate of 0.01.}
    \label{margin_conv}
\end{figure}

\begin{figure}
    \centering
\includegraphics[width=0.5\textwidth]{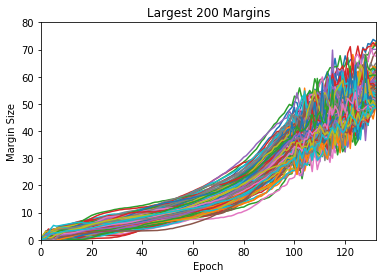}    
    \caption{Visualization of the largest 200 datapoint margins during training.}
    \label{margin_smallest}
\end{figure}

\begin{figure}
    \centering
\includegraphics[width=0.5\textwidth]{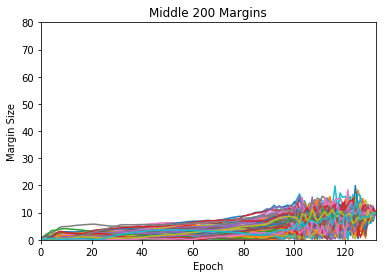}    
    \caption{Visualization of the middle 200 datapoint margins during training. }
    \label{margin_middle}
\end{figure}

\begin{figure}
    \centering
\includegraphics[width=0.5\textwidth]{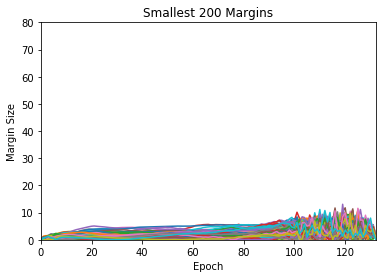}    
    \caption{Visualization of the smallest 200 datapoint margins during training. }
    \label{margin_largest}
\end{figure}

Figure \ref{margin_conv} shows the margins of 600 datapoints (200 smallest, 200 in the middle, and the 200 largest ones) throughout training. We note that prior to data separation, it is not feasible to predict which datapoints will have the smallest margins, as there is no indication of future margin performance from the margin information solely. Figures \ref{margin_smallest}, \ref{margin_middle}, and \ref{margin_largest} show the individual distributions of the smallest, middle, and largest 200 margins respectively, further reinforcing this observation. For all of these figures, we trained our 5-layer CNN on CIFAR10 with a batch size of 1, standard SGD, and a learning rate of 0.01. We note that data separation occurs around epoch 60.


\end{document}